\pdfoutput=1
\documentclass[11pt]{article}

\usepackage[]{EMNLP2023}

\usepackage{times}
\usepackage{latexsym}

\usepackage[T1]{fontenc}
\usepackage[utf8]{inputenc}

\usepackage{microtype}

\usepackage{inconsolata}
\usepackage{babel}
\usepackage{url}
\usepackage{color}
\usepackage{makecell}
\usepackage{bbm}
\usepackage{amsmath}
\usepackage{arydshln}
\usepackage{subcaption}
\usepackage{caption}
\usepackage{multirow}
\usepackage{graphicx}
\usepackage{bm}
\usepackage{soul}
\usepackage{amsthm}
\usepackage{xspace}
\usepackage{stmaryrd}

\newcommand{\smoe}{SMoE\xspace}
\newcommand{\spbleu}{BLEU\xspace}
\newcommand{\rc}{RC\xspace}
\newcommand{\rcs}{RCs\xspace}

\newcommand{\stratum}{stratum\xspace}
\newcommand{\strata}{strata\xspace}
\DeclareMathOperator\gate{Gate}
\DeclareMathOperator\ffn{FFN}
\newcommand{\config}[1]{\texttt{\smoe-#1}}

\usepackage{array, booktabs}

\usepackage{tikz}
\usepackage{xparse}

\definecolor{dimgray}{rgb}{.35,.35,.35}      % rgb(35, 35, 35)
\definecolor{darkgray}{rgb}{.20,.20,.20}     % rgb(20, 20, 20)

\definecolor{darkblue2}{HTML}{0614d6} % #0614d6 blue
\definecolor{darkred}{HTML}{a10303} % #a10303 red

\tikzset{
    txt/.style={
      rounded corners=1pt,
      text=darkgray,
      font=\small,
      align=center,
      inner sep=1pt,
      outer sep=2pt,
      minimum height=5mm,
    },
    conn/.style={
      -{Straight Barb[angle=60:1pt 2]}, line width=0.8pt,
        darkgray
    },
    expert/.style={
      fill=darkblue2!20,
      rectangle,draw=dimgray,
      rounded corners=2pt,
      text=darkgray,
      inner sep=2pt,
      minimum height=4mm,
      minimum width=.5cm,
      node distance=4pt and 10pt,
      line width=0.5pt,
      font=\small,
      align=center
    },
    gate/.style={
      fill=darkred!20,
      rectangle,draw=dimgray,
      rounded corners=2pt,
      text=darkgray,
      inner sep=2pt,
      minimum height=4mm,
      minimum width=.5cm,
      node distance=4pt and 10pt,
      line width=0.5pt,
      font=\small,
      align=center
    },
    lnorm/.style={
      fill=dimgray!20,
      rectangle,draw=dimgray,
      rounded corners=2pt,
      text=darkgray,
      inner sep=2pt,
      minimum height=4mm,
      minimum width=.5cm,
      node distance=4pt and 10pt,
      line width=0.5pt,
      font=\small,
      align=center
    },
    residual/.style={
        circle, draw=darkgray,minimum width=8pt,line width=0.8pt,
        path picture={
            \draw[darkgray]
            (path picture bounding box.south) -- (path picture bounding box.north) (path picture bounding box.west) -- (path picture bounding box.east);
    }}
}

\usetikzlibrary{decorations, shapes, arrows, arrows.meta}
\usetikzlibrary{positioning, decorations.pathreplacing, calc}
\usetikzlibrary{patterns}

\title{Towards Being Parameter-Efficient: A Stratified Sparsely Activated Transformer with Dynamic Capacity}

\author{Haoran Xu$^{\spadesuit}$, Maha Elbayad$^{\heartsuit}$, Kenton Murray$^{\spadesuit}$, Jean Maillard$^{\heartsuit}$, 
Vedanuj Goswami$^{\heartsuit}$\\ [1em]
$^{\spadesuit}$Johns Hopkins University, $^{\heartsuit}$Meta AI\\[1em]
\texttt{$\{$hxu64,kenton$\}$@jhu.edu}\\
\texttt{$\{$elbayadm,jeanm,vedanuj$\}$@meta.com}\\[1em]
}

\begin{document}
\maketitle
\begin{abstract}
Mixture-of-experts (MoE) models that employ sparse activation have demonstrated effectiveness in significantly increasing the number of parameters while maintaining low computational requirements per token.  However, recent studies \citep{hoffmannempirical,zuo2021taming,gao2022parameter} have established that MoE models are inherently \emph{parameter-inefficient} as the improvement in performance diminishes with an increasing number of experts. We hypothesize this parameter inefficiency is a result of  
all experts having equal capacity, which may not adequately meet the varying complexity requirements of different tokens or tasks. In light of this, we propose Stratified Mixture of Experts (\smoe) models, which feature a stratified structure and can assign dynamic capacity to different tokens. 
We demonstrate the effectiveness of \smoe on three multilingual machine translation benchmarks, containing 4, 15, and 94 language pairs, respectively. We show that \smoe outperforms multiple state-of-the-art MoE models with the same or fewer parameters.\footnote{Code is released at \url{https://github.com/fe1ixxu/Stratified_Mixture_of_Experts}.}
% On a diverse 15-language dataset, \smoe improves the translation quality over vanilla MoE by +0.93 \spbleu points on average. Additionally, \smoe is parameter-efficient, matching vanilla MoE performance with around 50\% fewer parameters.
\end{abstract}

\section{Introduction}
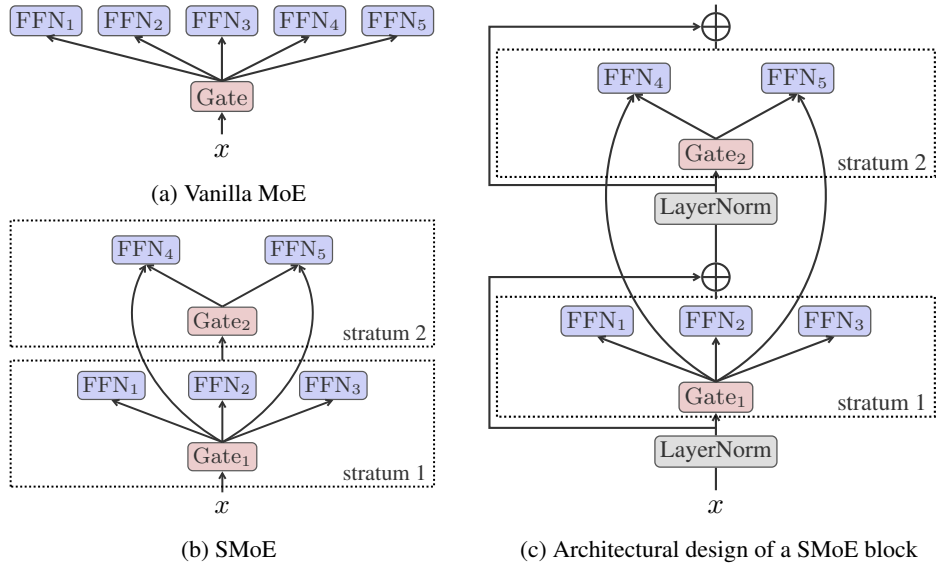
\begin{figure*}[ht]
         \centering
         %\resizebox{1\linewidth}{!}{
         %\includegraphics[width=\textwidth]{figures/moe_coombo.png}}
         \begin{tabular}[b]{c}
         \begin{subfigure}[b]{0.38\textwidth}
             \centering
             \resizebox{.95\linewidth}{!}{%
             \begin{tikzpicture}
    \node (input) at (0, 0) {$x$};
    \node[gate, above=3mm of input] (gate) {$\gate$};
    \node[expert, above=6mm of gate] (e3) {$\ffn_3$};
    \node[expert, right=2mm of e3] (e4) {$\ffn_4$};
    \node[expert, right=2mm of e4] (e5) {$\ffn_5$};
    \node[expert, left=2mm of e3] (e2) {$\ffn_2$};
    \node[expert, left=2mm of e2] (e1) {$\ffn_1$};

    \draw[conn] (input) -- (gate);
    \draw[conn] (gate.north) -- (e1.south);
    \draw[conn] (gate.north) -- (e2.south);
    \draw[conn] (gate.north) -- (e3.south);
    \draw[conn] (gate.north) -- (e4.south);
    \draw[conn] (gate.north) -- (e5.south);
\end{tikzpicture}
             }
             \caption{Vanilla MoE}
             \label{fig:intro_moe}
         \end{subfigure}\\
         \begin{subfigure}[b]{0.38\textwidth}
             \centering
             \resizebox{.95\linewidth}{!}{%
             \begin{tikzpicture}
    \node (input) at (0, 0) {$x$};
    \node[gate, above=3mm of input] (gate1) {$\gate_1$};
    \node[expert, above=6mm of gate1] (e2) {$\ffn_2$};
    \node[expert, right=6mm of e2] (e3) {$\ffn_3$};
    \node[expert, left=6mm of e2] (e1) {$\ffn_1$};

    \node[gate, above=5mm of e2] (gate2) {$\gate_2$};
    \node[expert, above left=6mm and 1mm of gate2] (e4) {$\ffn_4$};
    \node[expert, above right=6mm and 1mm of gate2] (e5) {$\ffn_5$};

    \draw[conn] (input) -- (gate);
    \draw[conn] (gate1.north) -- (e1.south);
    \draw[conn] (gate1.north) -- (e2.south);
    \draw[conn] (gate1.north) -- (e3.south);

    % Straight arrows
    %\draw[conn] (gate1.north) -- (e4.south);
    %\draw[conn] (gate1.north) -- (e5.south);

    % curvy arrows
    \draw[conn] (gate1.north) to [out=30,in=-60,loop,looseness=1] (e5.south);
    \draw[conn] (gate1.north) to [out=150,in=-120,loop,looseness=1] (e4.south);

    \draw[conn] (gate2.north) -- (e4.south);
    \draw[conn] (gate2.north) -- (e5.south);

    \draw[conn] (0, 2.1) -- (gate2.south);

    % borders:
    \draw[line width=0.8pt, densely dotted] (-3, 0.3) rectangle (3, 2.1);
    \draw[line width=0.8pt, densely dotted] (-3, 2.3) rectangle (3, 4.1);

    \node[txt, anchor=east] at (3, 0.5) {\stratum 1};
    \node[txt, anchor=east] at (3, 2.5) {\stratum 2};

\end{tikzpicture}
             }
             \caption{\smoe}
             \label{fig:intro_smoe}
         \end{subfigure}
         \end{tabular}%
     % \hfill
     \begin{subfigure}[b]{0.4\textwidth}
         \centering
         \resizebox{.95\linewidth}{!}{%
         \begin{tikzpicture}
    \node (input) at (0, 0) {$x$};
    \node[lnorm, above=3mm of input] (ln1) {LayerNorm};
    \node[gate, above=3mm of ln1] (gate1) {$\gate_1$};
    \node[expert, above=6mm of gate1] (e2) {$\ffn_2$};
    \node[expert, right=6mm of e2] (e3) {$\ffn_3$};
    \node[expert, left=6mm of e2] (e1) {$\ffn_1$};
    \node[residual, above=12mm of gate1] (res1) {};
    
    \node[lnorm, above=5mm of res1] (ln2) {LayerNorm};
    \node[gate, above=3mm of ln2] (gate2) {$\gate_2$};
    \node[expert, above left=6mm and 1mm of gate2] (e4) {$\ffn_4$};
    \node[expert, above right=6mm and 1mm of gate2] (e5) {$\ffn_5$};
    \node[residual, above=13mm of gate2] (res2) {};

    \draw[conn] (input) --(ln1) --  (gate1);
    \draw[conn] (gate1.north) -- (e1.south);
    \draw[conn] (gate1.north) -- (e2.south);
    \draw[conn] (gate1.north) -- (e3.south);

    % Straight arrows
    %\draw[conn] (gate1.north) -- (e4.south);
    %\draw[conn] (gate1.north) -- (e5.south);

    % curvy arrows
    \draw[conn] (gate1.north) to [out=30,in=-60,loop,looseness=1] (e5.south);
    \draw[conn] (gate1.north) to [out=150,in=-120,loop,looseness=1] (e4.south);

    \draw[conn] (res1) --(ln2) --  (gate2);
    \draw[conn] (gate2.north) -- (e4.south);
    \draw[conn] (gate2.north) -- (e5.south);

    % borders:
    \draw[line width=0.8pt, densely dotted] (-2.9, 1.2) rectangle (2.9, 2.8);
    \draw[line width=0.8pt, densely dotted] (-2.9, 4.4) rectangle (2.9, 6.1);

    \node[txt, anchor=east] at (2.9, 1.4) {\stratum 1};
    \node[txt, anchor=east] at (2.9, 4.6) {\stratum 2};

    % Residual connections:
    \draw[conn] ([yshift=-2mm]gate1.south) -- (
        [xshift=-3cm, yshift=-2mm]gate1.south) |- (res1.west);

    \draw[conn] ([yshift=-2mm]gate2.south) -- (
        [xshift=-3cm, yshift=-2mm]gate2.south) |- (res2.west);

    \draw[conn, -] (res1.south) -- ([yshift=-1mm]res1.south);
    \draw[conn, -] (res2.south) -- ([yshift=-1mm]res2.south);
    \draw[conn, -] (res2.north) -- ([yshift=1mm]res2.north);

\end{tikzpicture}
         }
         \caption{Architectural design of a \smoe block}
         \label{fig:smoe}
     \end{subfigure}
     \hfill
     \caption{ A high-level illustration of the vanilla MoE and \smoe. a) Vanilla MoE: gate is connected to all experts and sends tokens to top-$k$ selection.
     b) \smoe: Experts are stratified into $L$ \strata ($L{=}2$ in this example).
     Each \stratum has a gate that is connected to all subsequent experts.
     Tokens can be directly sent to the last \stratum to only experience one expert, or be sent to both \strata and have more capacity. Hence, the dynamic capacity of a token depends on how many experts it needs to pass. c) A detailed architectural design, where a comprehensive explanation of the design components will be presented in Section \ref{sec:method}.
     }
     \label{fig:intro}
\end{figure*}

Scaling up the model and data size has shown tremendous success in enhancing model performance across a large number of NLP tasks \citep{devlin-etal-2019-bert,conneau-etal-2020-unsupervised,kaplan2020scaling,gpt3}. Sparsely gated mixture of experts (MoE) \citep{shazeer2017,gshard} provides an effective way to greatly scale the model size under the same computational cost and achieves state-of-the-art performances on various tasks including natural language understanding \citep{switchtransformer}, machine translation \citep{nllb}, language modeling \citep{du2022glam}, etc. The efficiency comes from sparsely activating a subset of the neural network weights for each incoming sample. However, MoE is reported to be \textbf{parameter-inefficient} \citep{hoffmannempirical,zuo2021taming,gao2022parameter} i.e., there are diminishing improvement returns from adding more experts. For example, Switch Transformer \citep{switchtransformer} only outperforms T5 \citep{raffel2020exploring} by an average of 0.7 on the GLUE benchmark \citep{wang-etal-2018-glue} despite being 35$\times$ larger. Similarly, in the translation task, a MoE model with 20 times more parameters only offers an average improvement of 0.3 \spbleu on its ablation dataset (MoE-64 vs. 1.3B dense) \citep{nllb}.

We hypothesize that this parameter inefficiency stems from the equal capacity assignment, where we particularly define `\textbf{capacity}' as the \textbf{number of parameters used for the incoming token}.  For the current MoE models, the capacity of experts are the same used for serving all tokens. However, different tokens may demand varying capacities. For instance, in the context of multilingual machine translation, certain translation directions may necessitate a greater capacity to prevent overfitting, while others only require a smaller capacity.
To address this limitation, our hypothesis posits that the dynamic allocation of capacity to tokens results in more efficient utilization of parameters. Thus, we propose Stratified Mixture of Experts (\smoe) models, characterized by a stratified structure, which allows for the dynamic assignment of capacity to incoming tokens.

% We hypothesize that the parameter inefficiency could be caused by the static capacity assignment to all tokens, i.e., FFN layers that are assigned to handle incoming tokens always have the same capacity (the same number of parameters and architecture). However, the capacity requirements of different tokens may vary.
% For instance, some ambiguous words, such as "bank," which can refer to a financial institution or a river edge, may require more capacity to process their features, while other non-ambiguous words may not.
% Similarly, in the context of multilingual machine translation, certain translation directions may require a higher capacity to prevent overfitting, while others may only need a smaller capacity \maha{better to say low/high resource instead of certain directions}.
% To address this issue, we propose \textbf{Stratified Mixture of Experts (\smoe)} models with a hierarchical structure that can dynamically assign capacity to incoming tokens.

A high-level comparison of vanilla MoE and \smoe is presented in Figures \ref{fig:intro_moe} and \ref{fig:intro_smoe}.
In vanilla MoE, a single routing gate connects to all $E$ experts and sends tokens to the top-$k$ experts. Here, we take $E{=}5$ as an example.
In \smoe, the experts are divided into two \strata. Each \stratum has its own routing gate 
that connects to all experts in the current \stratum as well as all experts in the subsequent \strata.
If $\gate_1$ assigns tokens to Expert 4 or 5, the tokens will only need to pass through a single expert (an FFN layer). However, if tokens are sent to experts in the first \stratum (Experts 1 to 3), they will need to go through the next \stratum as well, meaning that another expert will be assigned by $\gate_2$ before exiting the \smoe block. 
This allows \smoe to dynamically assign capacity to different tokens. In addition, a comprehensive illustration of the architectural design of the SMoE model is provided in Figure \ref{fig:smoe}. A thorough explanation of the design elements will be provided in Section \ref{sec:method}.  Our main contributions are summarized as follows:\begin{itemize}
\itemsep0em
\item  We introduce the concept of \textbf{dynamic capacity} for MoE models and propose a mixture-of-experts model with a stratified structure, namely \smoe, which can automatically assign dynamic capacity to different incoming tokens to make experts become more parameter-efficient. 
\item We focus on the task of multilingual machine translation (MMT) and show that \smoe substantially outperforms numerous strong baselines with fewer than or the same number of parameters. For instance, we demonstrate that \smoe only needs half the number of parameters to achieve a performance on-par with a naive MoE \citep{gshard}. Furthermore, we carry out an in-depth analysis to probe the factors that impact dynamic capacity assignment, including the language of tokens and the position of the \smoe block within the model's architecture.
\end{itemize}
\section{Background and Related Work}
Massively multilingual machine translation models have been developed to handle several translation directions simultaneously in a single model  \citep{aharoni-etal-2019-massively}. However, the use of shared parameters for different languages often leads to negative transfer and decreased performance \citep{conneau-etal-2020-unsupervised,fan2020beyond}. In contrast to dense MMT models, sparsely gated mixture-of-experts (MoE) models, which activate a subset of parameters for each input, have been shown to significantly improve translation performance \citep{kim2021scalable,nllb}. \citet{shazeer2017} first demonstrated the benefit of adding MoE layers to scale RNN models for improved translation performance, and \citet{gshard} extended this work to transformer architectures \citep{transformer}.
MoE layers in the transformer model replace a single feedforward network (FFN) layer with $E$ FFN layers, denoted with $\{\text{FFN}_1, \ldots, \text{FFN}_E\}$. 
Each FFN layer is an expert.
Given an input token $x$, we have
\begin{align}
&\forall e\in \{1,\ldots,E\},\nonumber\\
&\text{FFN}_e(x) = W_1^{(e)}\text{ReLU}(W_2^{(e)}\cdot x),
\label{eq:ffn}
\end{align}
where $ W_1^{(e)}$ and $ W_2^{(e)}$ are the weights of $\text{FFN}_e$.
A trainable routing gate with weights $W_g$ predicts scores for these experts to use for the input $x$, in the form of a routing vector $G\in\mathbbm{R}^E$:
\begin{equation}
    G = \text{softmax}(W_g\cdot x).
    \label{eq:gate_score_naive}
\end{equation}
We select the set of top-$K$ experts, denoted with $\mathcal{E}\subset\{1,\cdots,E\}$, and compute the output of the MoE layer as follows:
\begin{equation}
    x_{\text{out}} = \sum\limits_{e\in\mathcal{E}}G_{e}\cdot\text{FFN}_e(x).
    \label{eq:expert_ffn_naive}
\end{equation}

% ---- Related Work
MoE models suffer from the notorious load imbalance issue, where the gate weights could collapse and send most tokens to the same expert. As a result, recent research has focused on designing better auxiliary load balancing loss functions to encourage tokens to be evenly distributed across experts, e.g., \citet{baselayer}  formulated token-to-expert allocation as a linear assignment problem, \citet{hashlayer} modified the feedforward layer to hash to different sets of weights depending on the current token, \citet{zoph2022designing} proposed a router z-loss that resolves instability issues, and \citet{zhou2022mixture} reversely design an expert-to-token allocation algorithm. Other lines of investigation in MoE include regularization techniques such as gating dropout \citep{liu2022gating} and output masking (EOM and FOM) \citep{nllb},
as well as novel MoE architectures, such as conditional MoE routing (CMR) that add an extra branch beside MoE layer \citep{nllb}, or Pyramid-Residual MoE~\citep{rajbhandari2022deepspeed}, a hybrid dense and MoE model with more experts in the last layers.

However, all previous work default to equal capacity for all tokens regardless of their language, frequency, or any other property. 
In the subsequent sections, we present a Stratified  Mixture of Experts (\smoe) model that automatically assigns dynamic capacities to different types of tokens.
% \section{Method}
% \citep{baselayer,hashlayer,kudugunta-etal-2021-beyond-distillation,switchtransformer,zoph2022designing,zhou2022mixture},
\section{Stratified Mixture of Experts}
\label{sec:method}
% \subsection{Hierarchical Mixture of Experts}
\subsection{Architectural Design}
The guiding design principle for Stratified Mixture of Experts (\smoe) is to assign dynamic capacity to tokens.
Given $E$ experts in an MoE block, we partition them into $L$ \strata.  The $i^{th}$ \stratum has a gate $\gate_i$ which routes tokens to an expert in the current \stratum as well as the subsequent ones.
This means that tokens can never be sent back to the previous \strata. 
Tokens keep getting routed in the \smoe block until they reach the final \stratum.
Different tokens will pass through a varying number of experts, resulting in different capacities, according to the assignment of gates. In vanilla MoE, however, the capacity through which every token goes is that of a single FFN layer. The workflow of how a token passes a \smoe block is shown in Figure \ref{fig:smoe}. For example, some tokens in the $1^{st}$ \stratum may be assigned to experts in the $2^{nd}$ \stratum while others are sent to the $3^{rd}$ \stratum. After several rounds of assignments, tokens finally exit the block after reaching the last \stratum. 

In \smoe, the successive application of multiple FFN layers to a token can result in training instability. Therefore, following the approach of \citet{transformer,wang2019learning,xiong2020layer}, we incorporate layer normalization~(LayerNorm, \citet{ba2016layer})
before dispatching tokens to experts and a residual connection after the tokens have passed through the experts. See Figure \ref{fig:smoe} for an overview of the design of our stratified experts.

% \begin{figure}[ht]
%     \centering
%     \resizebox{1\linewidth}{!}{
%     \includegraphics[width=7.5cm]{figures/hmoe.png}}
%     \caption{Workflow of how a token passes an \smoe block. The representation of the token is firstly layer normalized. Then, it is dispatched to the top-$k$ ($k=2$) experts, where the output is a linearly weighted combination of $k$ experts. Note that the new location of the token is the expert which has the highest score. Finally, a residue is added to the output of experts. The token will repeatedly pass layers of \smoe until it hits the last layer to exit. }
%     \label{fig:hmoe}
% \end{figure}

Formally, given $T$ tokens in a mini-batch, 
% $\{x_1, x_2, \cdots. x_T\}$,
we denote the $d$-dimensional representation of the $t^{th}$ token in the $i^{th}$ \stratum of the current \smoe block with $x_{i,t}$. Let $\mathcal{E}_i$ be the set of experts visible to the current gate (current \stratum plus subsequent \strata) and let $E_i=|\mathcal{E}_i|$ be its cardinality. 
% The gate weight in the $i^{th}$ \stratum is $(W_i)_{|E_i|\times d}$.
Before being dispatched to FFN layers, tokens are firstly normalized with LayerNorm,
\begin{equation}
    x_{i,t}' = \text{LayerNorm}(x_{i,t}).
    \label{eq:layer_norm}
\end{equation}
 Then, $\gate_i$ with weights $W_i$ predicts a probability distribution $G_{t}\in\mathbbm{R}^{E_i}$,
 scoring all visible $E_i$ experts at that \stratum:
\begin{equation}
   G_{t} = \gate_i(x_{i, t}') = \text{softmax}(W_i\cdot x_{i,t}').
    \label{eq:gate_score}
\end{equation}
% The top-$k$ values are selected for routing token $x_{i,t}$. The highest score determines the layer location for the next round of token assignment.
Following \citet{gshard}, we dispatch each token to at most $k{=}2$ experts. 
If $\mathcal{E}$ is the set of selected top-$k$ experts and the expert with the highest score is in the $j^{th}$ ($j>i$) layer, $x_{i,t}$ will be assigned to the $j^{th}$ layer and the output computation on the token is:
\begin{equation}
    x_{j,t} = \sum_{e\in\mathcal{E}}G_{t,e}\text{FFN}_e(x_{i,t}'),
    \label{eq:expert_ffn}
\end{equation}
where $G_{t,e}$ is the gate score for the expert $e$. Finally, We employ a residual connection after the FFN layer:
\begin{equation}
    x_{j,t} = x_{j,t} + x_{i,t}.
    \label{eq:residual_connect}
\end{equation}
Tokens gradually pass experts in deeper layers until they finish passing the final layer.

\subsection{Load Balancing}
Similar to \citet{gshard},
% ,switchtransformer,nllb},
we encourage tokens to be uniformly distributed across all visible experts. Each gate has a loss term to balance the load. For $\gate_i$, the loss is:
\begin{equation}
    \mathcal{L}_i = E_i \sum_{e\in \mathcal{E}_i} f_ep_e,
    \label{eq:gate_loss_i}
\end{equation}
where $f_e$ is the fraction of tokens dispatched to expert $e$, as their first choice, through top-$k$-gating:
\begin{equation}
    f_e = \frac{1}{T}\sum_{t=1}^T\mathbbm{1}\{\arg\max G_{t,e} = e\},
    \label{eq:gate_loss_f}
\end{equation}
and $p_e$ is the average routing probability to that expert over the T tokens in
the mini-batch:
\begin{equation}
    p_e = \frac{1}{T}\sum_{t=1}^T G_{t,e}.
    \label{eq:gate_loss_p}
\end{equation}
The auxiliary loss for the current \smoe block is computed by taking the average of the loss over all gates within the block.
\begin{equation}
    \mathcal{L} = \alpha\cdot\frac{1}{L}\sum_{i=1}^{L}\mathcal{L}_i,
    \label{eq:gate_loss}
\end{equation}
where $\alpha$ is a hyperparameter to control the strength of the load balancing loss. We average the loss over all \smoe blocks in the architecture as the final auxiliary loss appended to the original task loss.

\section{Experiments}
\label{sec:experiments}
We evaluate the proposed \smoe on a \textit{many-to-many} multilingual neural machine translation task. 

\subsection{Datasets}
In this study, we consider three datasets comprising 4, 15, and 94 languages each. The initial two datasets are extracted from the primary bitexts of the NLLB-200 training dataset, and we adopt their resource-level categorizations: \textit{high-resource} ($\ge$ 1M), \textit{very low-resource} ($\le$ 100K), and low-resource (the remaining).\footnote{Contrary to \citet{nllb}, in our study, \textit{very low-resource} is not included in the \textit{low-resource} category but considered as an independent set.} These two datasets are developed and evaluated using the Flores-200 dataset. The third dataset is OPUS-100 \citep{improvingmmt}. We follow \citet{improvingmmt} and divide directions into \textit{high-resource}, \textit{medium-resource}, and \textit{low-resource} categories.

\paragraph{NLLB M4 dataset}
% We first evaluate the performance of the \smoe model on a dataset consisting 
From NLLB, we pick 4 languages from 4 different linguistic families, 2 of which are high-resource and the other 2 are low-resource: Northern Sotho (\texttt{nso}, 526K parallel sentences), Malay (\texttt{msa}, 1M), Tagalog (\texttt{tgl}, 1M), Catalan (\texttt{cat}, 634K), totaling 3.2M training examples.

\paragraph{NLLB M15 dataset}
 Taking into account linguistic diversity and larger data size, we expand 
 the M4 dataset to cover a set of diverse 15 languages.
 M15 covers 6 linguistic families and a balanced number of high-resource, low-resource, and very low-resource languages (each category has 5 languages). 
 We show a detailed listing and information on the M15 dataset in Appendix \ref{app:m15_info}.

\paragraph{OPUS-100}
In addition to the datasets derived from NLLB, we also utilize OPUS-100 to examine a scenario involving a larger number of languages. OPUS-100 encompasses a total of 100 languages, which supports 94 development/test language pairs.

\paragraph{Evaluation}
During inference, we use beam search with a beam size of 5 and a length penalty of 1.0. We report BLEU scores \citep{papineni-etal-2002-bleu} for models trained on NLLB dataset and sacrebleu \citep{post2018call} with \texttt{flores200} tokenizer for models trained on OPUS-100. 

\subsection{Baselines}
We use five strong baselines to evaluate the effectiveness of \smoe. 
All baselines are our own implementation following the settings from the original papers. Note that the total number of experts i.e., the full model capacity is kept constant for all models in order to ensure fair comparison (8 experts for M4 and 16 experts for M15).

\paragraph{Vanilla MoE.}
An MoE model with Top-2 gating~\citep{gshard}. 

\paragraph{Switch Transformer.}
% Switch transformer \citep{switchtransformer} is a MoE method that mitigates training instability and reduces communication and computational costs. It also simplifies the routing algorithm to top-1-gating. 
An MoE model with Top-1 gating~\citep{switchtransformer}. Switch Transformer was introduced to mitigate training instabilities and improve the efficiency of Top-2 MoE models. Note that switch transformer uses fewer FLOPs per token due to the top-1 gating approach.

\paragraph{MoE + EOM.}
A vanilla MoE model regularized with Expert Output Masking (EOM) \citep{nllb}. EOM masks the expert output for a random fraction ($p_{eom}$) of outputs. We set $p_{eom}=0.1$ following the suggestion of \citet{nllb}

\paragraph{Conditional MoE Routing (CMR).}
CMR~\citep{nllb} augments MoE with a binary gate that sends tokens to one of two branches: (1) a shared FFN layer and (2) an vanilla MoE layer. Note that this method requires extra parameters due to the added shared FFN layer. The CMR budget constraint is set to 0.8. 
% as suggested by \citet{nllb}. 

\paragraph{Stacking MoE Layers.}
A similar model architecture to \smoe where we simply stack multiple MoE layers, e.g., stacking 2 MoE layers with 4 experts each vs. one block of \smoe with 2 \strata where each \stratum has 4 experts. Unlike \smoe where each expert is surrounded with a residual skip connection and preceded with a LayerNorm, here the stacking is naive without any addition. 

\subsection{Training Details}
Following \citet{johnson-etal-2017-googles}, we prepend source sentences with a special language token \texttt{<2xxx>} to indicate the target language. 
We use a data sampling temperature of $T{=}1$ suggested by \citet{nllb} to train on NLLB datasets, and $T{=}5$ suggested by \citet{improvingmmt} to train on OPUS-100. 

The dense model architecture, backbone for all trained models, is a Transformer model~\citep{transformer} with 12 layers (6 on encoder and 6 on decoder). We use transformer$_\texttt{base}$ and $E{=}8$ experts for M4,  and transformer$_\texttt{big}$ and $E{=}16$ experts for M15 and OPUS-100.\footnote{transformer$_\texttt{base}$: FFN dimension of 2048, 8 heads, and embedding dimension of 512; transformer$_\texttt{big}$: FFN dimension 4096, 16 heads, and embedding dimension 1024.}
% The configurations we use for three datasets: 
% \begin{itemize}
% \item M4: FFN dimension of 2048, 8 heads, and embedding dimension of 512. The total number of experts per layer in MoE models is set to $E{=}8$.
% \item M15: FFN dimension 4096, 16 heads, and embedding dimension 1024. The total number of experts per layer in MoE models is set to $E{=}16$.
% \item OPUS-100: The same as M15.
% \end{itemize}

In MoE models, every other FFN layer of the encoder and decoder are substituted with an MoE layer.
For \smoe, in the $i^{th}$ \stratum, we enforce that each expert processes, at most, $2\times T_i/E_i$ tokens, where $T_i$ is the number of tokens in the mini-batch sent to the layer $i$ and $E_i$ is the number of visible experts. 
For the other MoE baselines, it is $2\times T/E$, where T is the number of tokens in the mini-batch and $E$ is the total number of experts. 
The multiplicative coefficient $\alpha$ for the auxiliary load balance loss is set to 0.01. 
A vocabulary of size 32k for both M4 and M15 and 64K for OPUS-100 with SentencePiece~\citep{sentencepiece}. For a fair comparison, All models are trained for the same number of updates. More details can be found in Appendix \ref{app:train_detail}.
%
% We train with a batch size of 4096 tokens per GPU and all models are trained on 32 Tesla-V100 up to 100K steps. The learning rate is warmed up over the first 8K steps to a peak value of 0.0008, from which the learning rate inverse square root decayed.

\begin{table*}[ht]
\centering
\resizebox{1\linewidth}{!}{
% \begin{tabular}{l|cccc:cccc:c|cc}
\begin{tabular}{lccccccccccc}
\toprule
% Methods                          & nso$\rightarrow$eng & msa$\rightarrow$eng & tgl$\rightarrow$eng & cat$\rightarrow$eng & eng$\rightarrow$nso & eng$\rightarrow$msa & eng$\rightarrow$tgl & eng$\rightarrow$cat & Avg.           & \# Parameters & FLOPs$/$tok \\ 
& \multicolumn{4}{c}{xxx$\rightarrow$eng} & \multicolumn{4}{c}{eng$\rightarrow$xxx} & \multirow{2}{*}{Avg.} & \multirow{2}{*}{\#Parameters} & \multirow{2}{*}{FLOPs$/$tok}\\
\cmidrule(lr){2-5} \cmidrule(lr){6-9}
% \cmidrule
Methods (8 experts) & nso & msa & tgl & cat & nso & msa & tgl & cat & & &  \\ 
\midrule
Dense Model \citep{transformer}                & 28.87                 & 37.01                 & 41.76                 & 40.35                 & 26.68                 & 44.30                 & 38.20                 & 45.50                 & 37.83          & 60M & 167M          \\
Vanilla MoE \citep{gshard}& 29.45                 & 41.26                 & 44.23                 & 42.91                 & 27.59                 & 47.12                 & 40.10                  & 47.66                 & 40.04          & 148M &192M          \\
Switch Transformer~\citep{switchtransformer}             & 29.69                 & 40.35                 & 44.50                 & 43.19                 & 27.73                 & 46.44                 & 39.88                 & 47.26                 & 39.88          & 148M  & 167M        \\
MoE+EOM \citep{nllb}                             & 30.52                 & 40.50                  & 44.52                 & 43.48                 & 27.71                 & 46.85                 & 40.09                 & 47.14                 & 40.10          & 148M  & 192M        \\
MoE+CMR \citep{nllb}                               & 30.54                 & 41.29                 & 44.65                 & 43.71                 & 27.76                 & 46.87                 & {\underline{40.36}}           & 47.65                 & 40.35          & 161M  & 217M       \\ \midrule
Stacking 2 4-expert MoE & 28.65          & 39.24          & 42.02          & 41.36          & 27.58          & 46.27          & 40.06          & 47.50          & 39.09          & 148M & 242M\\
\config{4-4}                         & 30.14                 & 41.05                 & \textbf{45.13}        & \textbf{43.97}        & \textbf{27.86}        & {\underline{47.48}}           & 39.85                 & {\underline{47.71}}           & {\underline{40.40}}    & 148M    & 217M      \\
Stacking 4 2-expert MoE & 25.53          & 35.53          & 38.34          & 37.43          & 24.99          & 41.35          & 36.85          & 42.15          & 35.27          & 148M & 327M \\
\config{2-2-2-2}                     & \textbf{30.96}        & \textbf{41.88}        & {\underline{44.86}}           & {\underline{43.37}}           & {\underline{27.83}}           & \textbf{47.96}        & \textbf{40.95}        & \textbf{48.39}        & \textbf{40.78} & 148M & 247M          \\
\bottomrule
\end{tabular}
}
\caption{Overall \spbleu results on the M4 dataset. The best values are bold and the second-best values are underlined. 
The number of experts is 8 for all methods. 
The two \smoe models attain the two best performances across all languages.} 
\label{tab:m4_results}
\end{table*}

\begin{table*}[ht]
\centering
\resizebox{1\linewidth}{!}{
% \begin{tabular}{lc|cccc:cccc:c|cc}
\begin{tabular}{llccccccccccc}
\toprule
% \multicolumn{2}{l|}{Methods}                               & \multicolumn{4}{c:}{eng$\rightarrow$xxx}                         & \multicolumn{4}{c:}{xxx$\rightarrow$eng}                         & Avg.           & \# Parameters & FLOPs$/$tok \\ 
&  & \multicolumn{4}{c}{eng$\rightarrow$xxx} & \multicolumn{4}{c}{xxx$\rightarrow$eng} & \multirow{2}{*}{Avg.} & \multirow{2}{*}{\#Parameters} & \multirow{2}{*}{FLOPs$/$tok} \\ 
\cmidrule(lr){3-6} \cmidrule(lr){7-10}
% \multicolumn{2}{l|}{}
\multicolumn{2}{l}{Methods (default is 16 experts)}  
& all & high & low & very low  & all & high & low & very low &  &   &  \\
\midrule
\multicolumn{2}{l}{Dense Model \citep{transformer}}                           & 29.27          & 38.68          & 32.44          & 16.69          & 29.94          & 37.55          & 32.37          & 19.89          & 29.61          & 209M      & 506M    \\
\multicolumn{2}{l}{Vanilla MoE~\citep{gshard}}      & 32.01          & 40.41          & 34.77          & 20.84          & 32.00          & 39.38          & 34.73          & 21.89          & 32.00          & 963M & 606M         \\
\multicolumn{2}{l}{Switch Transformer \citep{switchtransformer}}                    & 31.67          & 40.44          & 34.72          & 19.87          & 32.09          & 39.85          & 34.75          & 21.68          & 31.89          & 963M  & 506M        \\
\multicolumn{2}{l}{MoE+EOM~\citep{nllb}}                                   & 32.02          & 40.63          & 35.19          & 20.25          & 32.81          & {\underline{40.42}}    & 35.33          & 22.69          & 32.41          & 963M & 606M         \\
\multicolumn{2}{l}{MoE+CMR~\citep{nllb}}                                   & 31.83          & 40.22          & 34.81          & 20.46          & \underline{33.02}          & 40.41          & 35.42          & 23.21          & 32.42          & 1.01B   & 707M      \\ \midrule
\multirow{3}{*}{2-layer \smoe}           & \config{4-12}             & \textbf{33.00} & 41.14          & {\underline{35.77}}    & \textbf{22.10} & 32.86          & \textbf{40.63} & 35.14          & 22.80          & \textbf{32.93} & 963M & 656M         \\  
& \config{12-4}             & 32.52          & 41.11          & 35.56          & 20.89          & 32.94          & 40.03          & \textbf{35.21} & {\underline{23.60}}    & 32.73          & 963M   & 757M      \\
& \config{8-8}              & 32.28          & 41.24          & 35.33          & 20.27          & 32.72          & 40.30          & 34.80           & 23.08          & 32.50          & 963M & 707M          \\ 
\midrule
\multirow{2}{*}{3-layer \smoe} & \config{4-4-8}   & 32.83          & \textbf{41.61} & \textbf{35.82} & 21.07          & 32.24          & 39.40          & 34.53          & 22.80          & 32.54          & 963M   &  724M     \\
 & \config{8-4-4}   & 32.24          & 40.99          & 35.12          & 20.61          & \textbf{33.06} & 40.36          & \underline{35.12}    & \textbf{23.71} & 32.65          & 963M & 805M          \\ 
 \midrule
4-layer \smoe    & \config{4-4-4-4}          & {\underline{32.87}}    & {\underline{41.50}}    & 35.68          & {\underline{21.42}}    & 32.63          & 40.03          & 34.50          & 23.35          & {\underline{32.75}}    & 963M  & 825M  \\ 
\midrule
% \multicolumn{2}{l|}{Switch Transformer, 32 experts, top-2 \citep{switchtransformer}} 
\multicolumn{2}{l}{Vanilla MoE, 32 experts}
& 32.98          & 41.05          & 35.69          & 22.19          & 32.90          & 40.79          & 35.49          & 22.42          & 32.94          & 1.77B    & 606M     \\ 
\bottomrule
\end{tabular}
}
\caption{Overall \spbleu results on the M15 dataset. The best values are bold and the second-best values are underlined. 
% The number of experts designated as default is 16 if no other specifications.
Unless otherwise mentioned, the number of experts is 16.
All \smoe models outperform the baselines. The best setting is \config{4-12}, which outperforms vanilla MoE by +0.93 \spbleu. 
Vanilla MoE would require to double its parameters to achieve similar performance to \config{4-12}.}
\label{tab:m15_results}
\end{table*}

\begin{table*}[ht]
\centering
\resizebox{1\linewidth}{!}{
% \begin{tabular}{lc|cccc:cccc:c|cc}
\begin{tabular}{llccccccccc}
\hline
\multicolumn{2}{l}{}                        & \multicolumn{4}{c}{eng$\rightarrow$xxx} & \multicolumn{4}{c}{xxx$\rightarrow$eng} & \multirow{2}{*}{Avg.} \\
\cmidrule(lr){3-6} \cmidrule(lr){7-10}
\multicolumn{2}{l}{Methods}                 & all      & high    & medium   & low     & all      & high    & medium   & low     &                       \\ \hline
\multicolumn{2}{l}{Dense Model \citep{transformer}} & 27.37    & 23.89   & 31.17    & 29.76   & 30.60    & 29.40   & 31.85    & 31.49   & 28.99                 \\
% \multicolumn{2}{l}{CLSR \citep{zhang2021share}}                    & 27.39    & 23.91   & 31.17    & 29.77   & 32.43    & 30.73   & \textbf{33.85}    & \textbf{34.20}   & 29.91                 \\
\multicolumn{2}{l}{Vanilla MoE \citep{gshard}}             & 30.34    & 26.16   & 34.78    & 33.38   & 32.38    & 31.20   & 33.67    & 33.21   & 31.36                 \\
\multicolumn{2}{l}{Switch Transformer \citep{switchtransformer}}      & 30.03    & 25.76   & 34.46    & 33.27   & 31.44    & 30.58   & 33.00    & 31.18   & 30.74                 \\
\multicolumn{2}{l}{MoE+EOM \citep{nllb}}                 & 30.48    & 26.24   & 34.76    & 33.86   & 32.45    & 31.23   & 33.69    & \underline{33.44}   & 31.47                 \\
\multicolumn{2}{l}{MoE+CMR \citep{nllb}}                     & 30.56    & 26.32   & 34.98    & 33.77   & 32.43    & 31.33   & 33.69    & 33.11   & 31.50                 \\ \hline
\multicolumn{2}{l}{\config{4-12}}               & \textbf{32.15}    & \textbf{27.99}   & \textbf{36.67}    & \underline{35.06}   & \underline{32.58}    & \textbf{32.42}   & 33.76    & 31.37   & \textbf{32.37}                 \\
\multicolumn{2}{l}{\config{4-4-4-4}}            & \underline{31.84}    & \underline{27.31}   & \underline{36.52}    & \textbf{35.33}   & \textbf{32.71}    & \underline{31.87}   & \underline{33.80}    & 33.04   & \underline{32.28}                 \\ \hline
\end{tabular}
}
\caption{Overall \spbleu results on the OPUS-100 dataset. The best values are bold and the second-best values are underlined. The number of experts is 16. We consider the two best settings in M15 dataset, \config{4-12} and \config{4-4-4-4}. Both of them substantially outperform all baselines. The number of parameters and FLOPs/tok for MoE models are the same as Table \ref{tab:m15_results}.}
\label{tab:opus_results}
\end{table*}

\subsection{\smoe configurations}
We use a series of numbers separated by hyphens
to describe the \smoe configuration. For instance, \config{4-4-8} indicates that all MoE blocks have 3 \strata,
where the $1^{st}$ \stratum has 4 experts, the $2^{nd}$ has 4 and the $3^{rd}$ has 8.
% For M4, we consider 2 configurations of \smoe: \config{4-4} and \config{2-2-2-2}. 
% For M15, we consider a multitude of settings splitting the 16 experts per layer over 2, 3 or 4 \strata.
% % the settings are more diverse and complicated. We consider \smoe ranging from 2 layers to 4 layers. 
% \smoe does not necessarily need to be balanced, i.e., have an unequal number of experts per \stratum. We refer to the first column of Table \ref{tab:m15_results} to see the list of selected configurations.

\subsection{Results}
\paragraph{M4 Results}
The results are in Table \ref{tab:m4_results}. We consider two configurations for \smoe: \config{4-4} and \config{2-2-2-2}. The better \smoe settings for M4 is \config{2-2-2-2}. In M4, \config{2-2-2-2} outperforms Switch Transformer by +0.9 \spbleu on average and vanilla MoE by +0.74 \spbleu. Out of 5 MoE baselines, CMR achieves the best performance (+0.4 \spbleu over Switch Transformer and +0.3 \spbleu over vanilla MoE), however, CMR models have more parameters as well as more FLOPs per token.

 It is worth noting that simply stacking MoE layers degenerates the model performance, which indicates the importance and effectiveness of the specific design of \smoe.
 % We omit experiments of stacking MoE in M15 due to the large performance drop in M4. 
\begin{figure*}[ht]
     \centering
     \begin{subfigure}[b]{0.5\textwidth}
         \centering
         \resizebox{1\linewidth}{!}{
         \includegraphics[width=\textwidth]{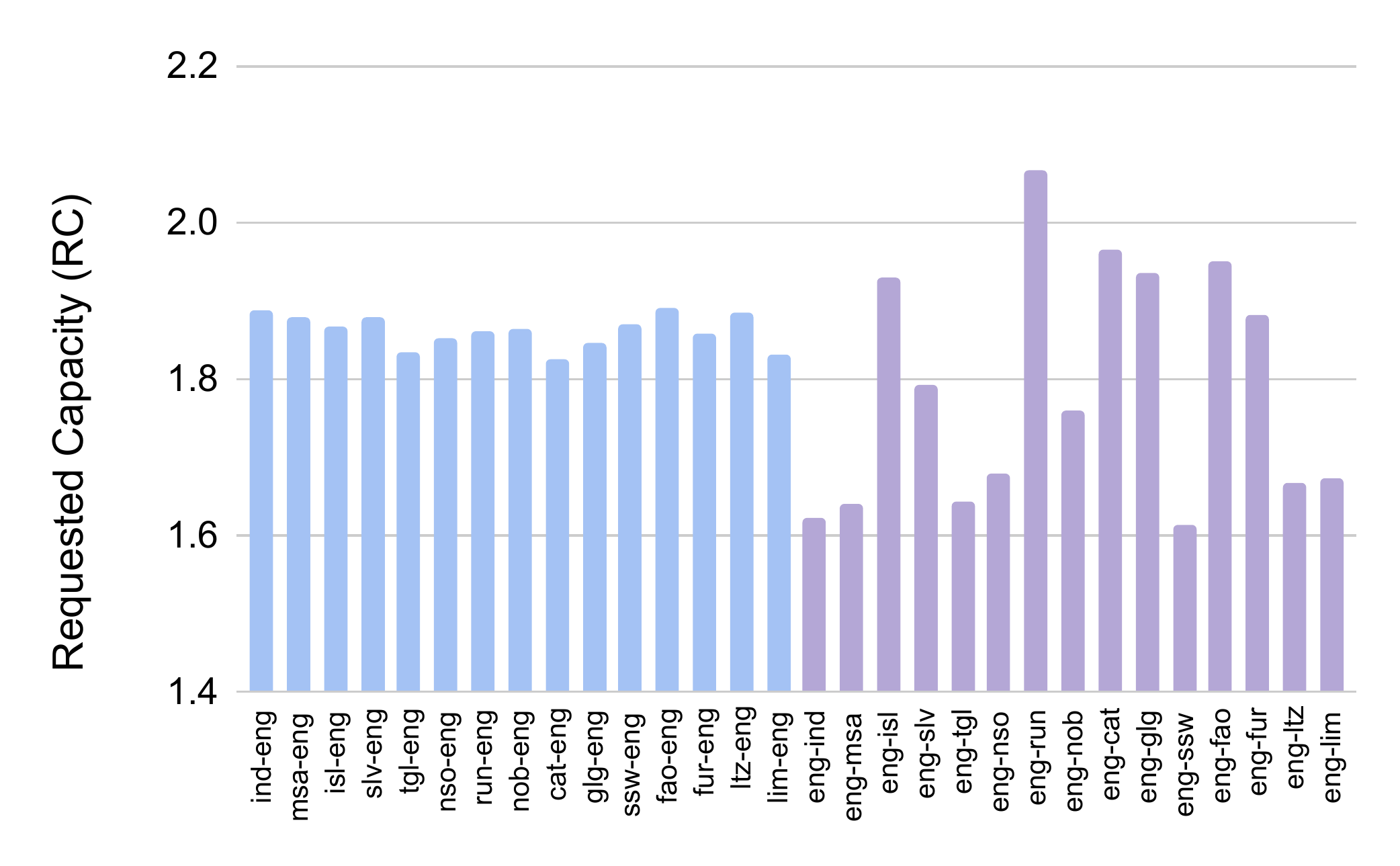}}
         \caption{Decoder}
         \label{fig:noc_lg_decoder}
     \end{subfigure}\hfill
     \begin{subfigure}[b]{0.5\textwidth}
         \centering
         \resizebox{1\linewidth}{!}{
         \includegraphics[width=\textwidth]{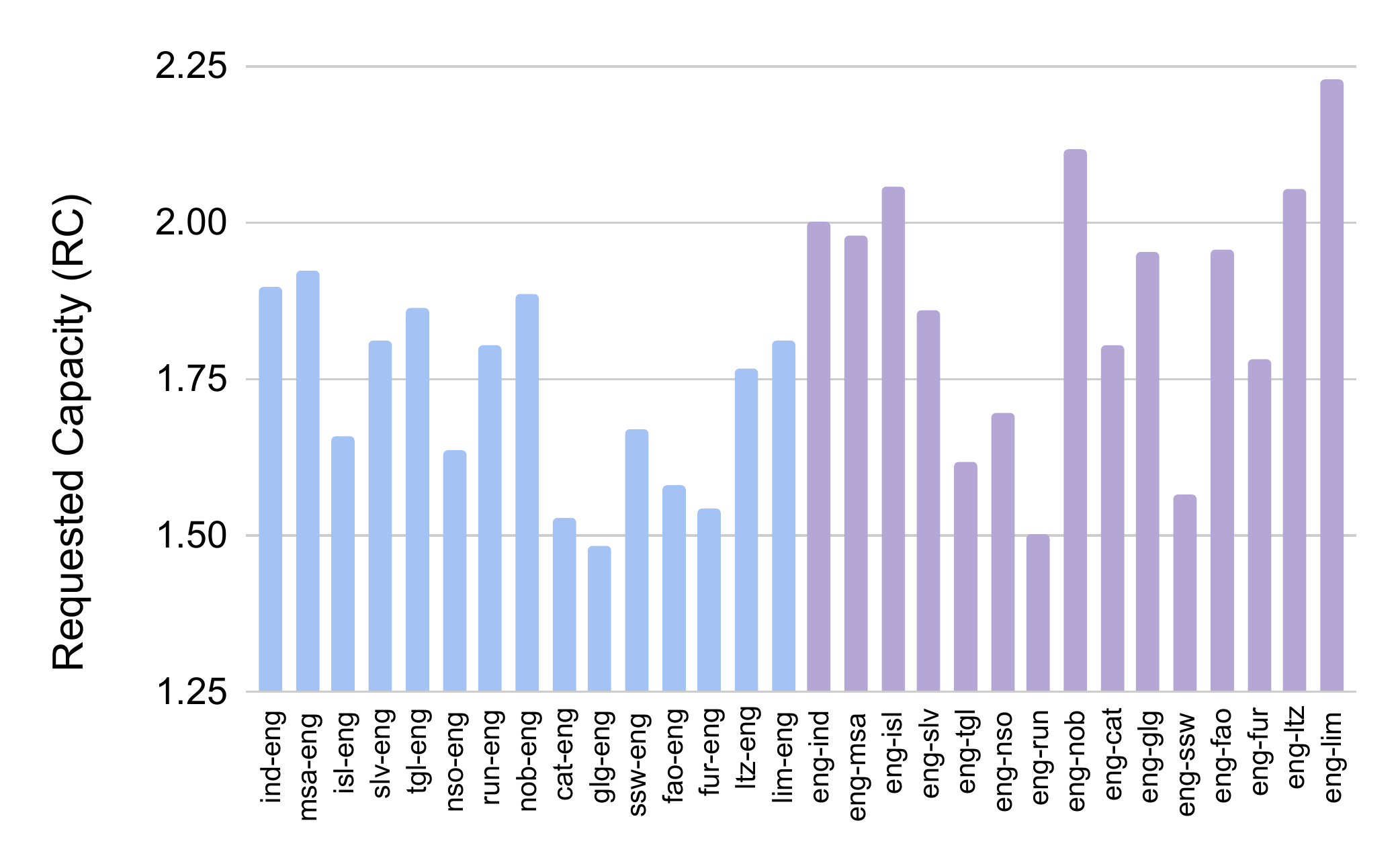}}
        \caption{Encoder}
         \label{fig:noc_lg_encoder}
     \end{subfigure}
     \caption{Average requested capacity (\rc) of all tokens in each translation direction. The blue bars are for \texttt{xxx}$\shortrightarrow$\texttt{eng} directions and the purple bars are for \texttt{eng}$\shortrightarrow$\texttt{xxx} directions. 
     Directions in each subset are sorted from high-resource to low-resource.
     On the decoder side, the average \rc of \texttt{eng} tokens is similar regardless of the source language,
     but averaged \rc has a large variance if the target language is different.
     On the encoder side, \rc is always different even though the source language is the same. 
     % One hypothesis is that the special symbol at the beginning of the source sentence can affect the capacity assignment. Overall, the \rc is sensitive to to the target language in the decoder and to the translation direction in the encoder.
     } 
     \label{fig:noc_lg}
\end{figure*}

\paragraph{M15 results}
We show results in Table \ref{tab:m15_results}. We consider a multitude of settings splitting the 16 experts per layer over 2, 3 or 4 \strata. The best \smoe settings for the larger M15 dataset is \config{4-12}. This configuration demonstrated an average improvement of +1.04 \spbleu over the Switch Transformer and +0.93 \spbleu over the vanilla MoE across the 15 languages evaluated. However, MoE+EOM and CMR only improve vanilla MoE by +0.52 and +0.53 \spbleu, respectively. Note that if vanilla MoE wants to achieve similar performance to \config{4-12} (32.94 vs. 32.93 \spbleu on average), it has to increase experts from 16 to 32, which almost doubles the total number of parameters from 963M to 1.77B (last row of Table \ref{tab:m4_results}), which means our model is much more parameter-efficient. 

\smoe models with more \strata, allowing for more depth, do not guarantee better performance. A clear example is \config{4-12} and \config{4-4-4-4} in M15 (32.93 vs. 32.75 averaged \spbleu). However, for `balanced' \smoe (equal number of experts per \stratum), fewer experts per \stratum achieves better performance: \config{2-2-2-2} outperforms \config{4,4} (40.78 vs. 40.40 \spbleu) on M4 and \config{4-4-4-4} outperforms \config{8,8} (32.75 vs. 32.50 \spbleu) on M15.

% Interestingly, for `unbalanced' \smoe, having more experts in the upper \strata of \smoe is beneficial to \texttt{eng}$\shortrightarrow$\texttt{xxx} direction 
% but may hurt the \texttt{xxx}$\shortrightarrow$\texttt{eng} performance, and vice versa.
% For instance, in Table \ref{tab:m15_results}, \config{4-12} obtains better performance than \config{12-4} on \texttt{eng}$\shortrightarrow$\texttt{xxx} (33.00 vs. 32.52) while worse on \texttt{xxx}$\shortrightarrow$\texttt{eng} (32.86 vs. 32.94). A similar trend can be observed with 3-layer \smoe. \config{4-4-8} outperforms \config{8-4-4} by +0.59 \spbleu on \texttt{eng}$\shortrightarrow$\texttt{xxx} but underperforms \config{8-4-4} by 0.82 \spbleu on \texttt{xxx}$\shortrightarrow$\texttt{eng}. Finding an optimal trade-off between \texttt{xxx}$\shortrightarrow$\texttt{eng} and \texttt{eng}$\shortrightarrow$\texttt{xxx} needs more empirical studies and we leave this work in the future.
\paragraph{OPUS-100 Results: A Larger Performance Gap.}
Table \ref{tab:opus_results} presents the comprehensive results. Notably, the performance disparity becomes more pronounced we scale our experiments to 94 languages ( OPUS-100 does not support the remaining 5 languages ). We select the two optimal \smoe configurations in M15: \config{4-12} and \config{4-4-4-4}. The \config{4-12} configuration consistently demonstrates superior performance, achieving a larger margin compared to our best baselines, EOM and CMR. The \config{4-12} outperforms our naive MoE model and Switch Transformer by +1.01 and +1.63 BLEU, significantly outperforming the gains achieved by EOM (+0.11 and +0.73) and CMR (+0.14 and 0.76). 
% Furthermore, we add CLSR \citep{zhang2021share} (CLSR + MoE is our CMR baseline) as a baseline in OPUS, given its dynamic parameter assignment (but fixed capacity), which is related to our dynamic capacity assignment.

% \textbf{Overall, \smoe outperforms all baselines with the same or less number of parameters.}
Overall, \smoe outperforms all baselines with the same or a fewer number of parameters.

\subsection{Computational Cost}
\label{sec:compute_cost}
% As a token may pass multiple experts in \smoe,
As a token may pass through multiple \strata in any given \smoe layer,
the average computational cost is higher than other MoE models. 
% given the same number of training updates.
In the last column of Tables \ref{tab:m4_results} and \ref{tab:m15_results}, we record FLOPs per token for all models, and Table \ref{tab:opus_results} shares the same information with Table \ref{tab:m15_results}.\footnote{FLOPs are calculated for the forward pass as done in \citet{kaplan2020scaling}.}
Although \smoe requires more FLOPs per token, the additional cost is only a marginal increase over vanilla MoE models.
For example, in M15 and OPUS-100, our best setting \config{4-12} merely uses 8\% more FLOPs/tok than other top-2-gating MoE models, 
but significantly outperforms all of them.
% We argue that the small extra cost is acceptable conditioned on the strong improvement.

\section{Analysis}
The advantage of \smoe is that it can assign dynamic capacity to different tokens, with some tokens passing through only one expert and others passing through multiple experts. 
Here, we define the \textbf{Requested Capacity (\rc)} as the average number of experts that a token need to pass in one \smoe block.
RC of a token is dictated by how the \smoe gates route it throughout the different \strata.
To understand what may affect the \rc of tokens, we examine three potential influencing factors: 
the language of the input token,
the frequency of the token, 
and the depth of the \smoe block. 
All analysis is conducted using the \config{4-4-4-4} model trained on the M15 dataset.

\subsection{The Language of The Input Token}
Here, we investigate whether different languages have different \rcs. 
We begin with collecting the average \rc in all translation directions for all tokens in the training and development (Flores-200 dev) sets.
We investigate \smoe blocks in the encoder and decoder separately as they process different tokens (source tokens for the encoder and target tokens for the decoder).
% , because the encoder and decoder receive different languages\footnote{For example, in \texttt{msa}$\shortrightarrow$\texttt{eng}, \smoe blocks in the encoder only receive \texttt{msa} tokens but decoder also receive \texttt{eng} tokens.}.
The average \rc is then averaged across all \smoe blocks in either encoder or decoder.

Figure \ref{fig:noc_lg_decoder} shows the average \rc in the decoder for each translation direction in M15.
When translating into English (\texttt{xxx}$\shortrightarrow$\texttt{eng}, blue bars), we observe that the target English tokens have similar \rc in the decoder side ($\approx$1.85 experts) irrespective of the source language.
When translating from English (\texttt{eng}$\shortrightarrow$\texttt{xxx}), the \rc varies a lot with respect to the target language.

Unlike in the decoder where only the target language matters, Figure \ref{fig:noc_lg_encoder} shows variability in \rc with respect to both source and target languages i.e, not only the language of the tokens themselves (source), but also the target language we will be translating into once we move to the decoder. 
% to the decoder, the difference of NoC is large if the input languages are different (\texttt{xxx}$\shortrightarrow$\texttt{eng}) in the encoder, as shown in . 
% However, differing from the decoder, \smoe also assigns \texttt{eng} tokens with various NoC when they have different target languages (\texttt{eng}$\shortrightarrow$\texttt{xxx}).
We hypothesize the special symbol at the beginning of the source sequence (\texttt{<2xxx>}) can affect the capacity assignment. 
In conclusion, capacity assignment is sensitive to the target language in the decoder and to the translation direction in the encoder.
% the language (or at least, translation direction), however, it could be affected by the special target language token.

\subsection{Token Frequency}
\begin{figure}[t]
    \centering
    \resizebox{1\linewidth}{!}{
    \includegraphics[width=7.5cm]{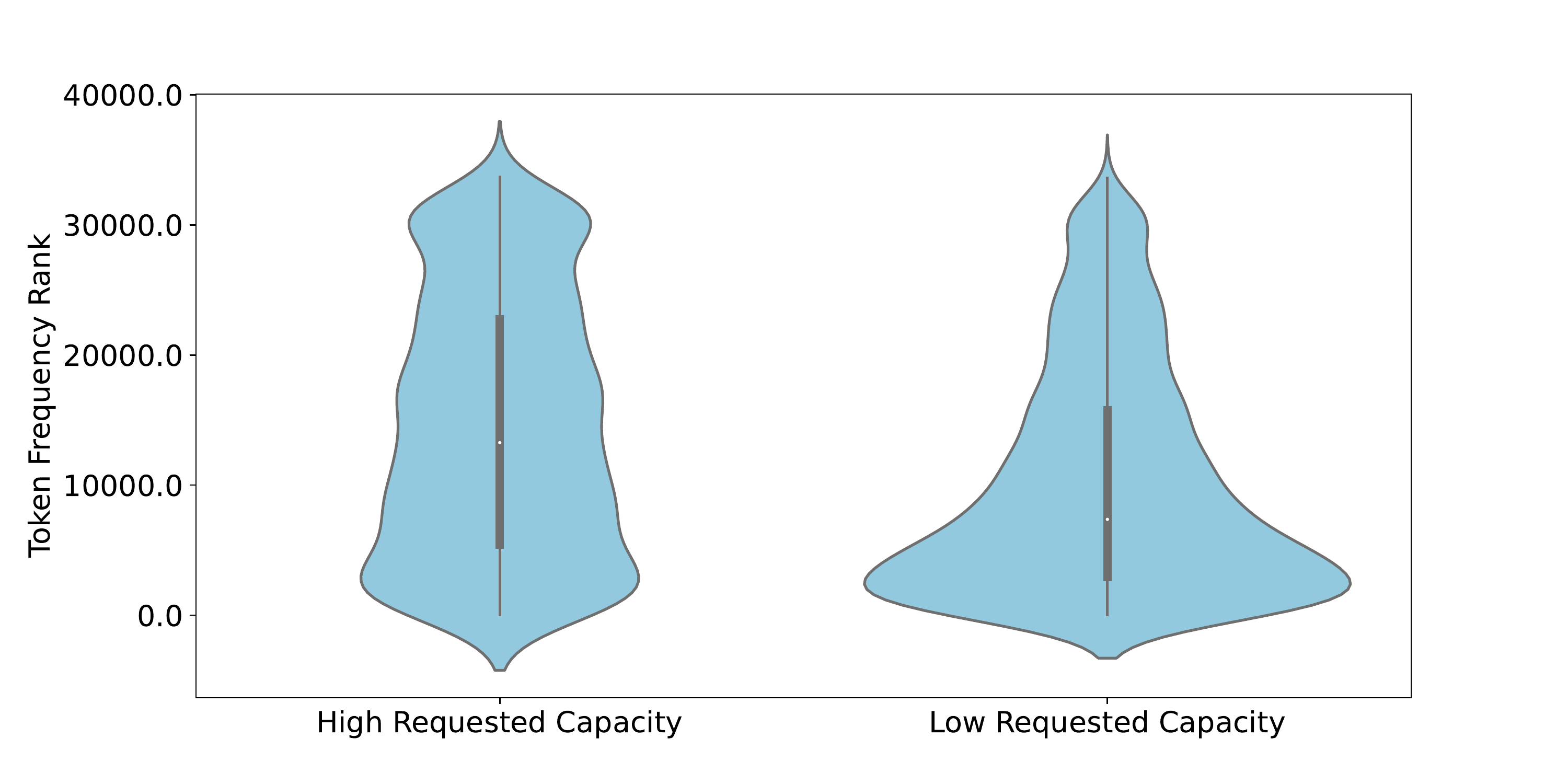}}
    \caption{
    Violin plots of the token frequency in high-\rc (left) and low-\rc (right) tokens.
    % We collect tokens that use the highest and lowest \rc and use violin plots to visualize the distribution of token frequency. 
    Unlike high-\rc tokens, low-\rc tokens tend to be highly frequent ones.
    }
    \label{fig:freq}
\end{figure}

As in the previous section, we record the average \rc for all tokens in training and development data,
% in our development set\maha{double check}
and in all translation directions.
To avoid looking at all 32K tokens in our vocabulary, we select the top-25 tokens with the highest \rc in each \smoe block and in each translation direction, totaling 4500 tokens.\footnote{25 $\times$ 6 (\#\smoe blocks) $\times$ 30 (\#directions) = 4500.}
We similarly collect the bottom-25 tokens with the lowest \rc. After removing tokens repeatedly selected by different directions or by different \smoe blocks, 
we end up with 2580 unique high-\rc tokens and 3208 unique low-\rc tokens.
% we totally collect 2580 tokens that used the highest capacity and 3208 tokens that used the lowest capacity. 
% Recall that the total number of types is 32000.
We draw in Figure \ref{fig:freq} a violin plot to show the distribution of tokens in these two groups in terms of their frequency in the training data.
We rank the frequencies on the y-axis so that a lower rank means more frequent tokens, e.g., rank 0 corresponds to the most frequent token in our training data.
The results show that there is no strong correlation between frequency and \rc for tokens with the highest \rc.
On the other end of the spectrum, tokens with the lowest \rc tend to be high-frequency tokens,
as indicated by the right violin plot being wider at the bottom part (rank < 10k). 
Many of these high-frequency tokens are basic subword units (like \texttt{\_li}, \texttt{\_el}, \texttt{\_pa}) or punctuation marks. One can interpret \rc as a metric for `difficulty in processing a given token'.
The model was overly exposed to these frequent tokens, and as such, does not require a lot of capacity to process them.
% Punctuation and common subword units are not considered to have a significant impact on the processing capacity required.

\subsection{Location of The \smoe Block}
\begin{figure}[t]
    \centering
    \resizebox{1\linewidth}{!}{
    \includegraphics[width=7.5cm]{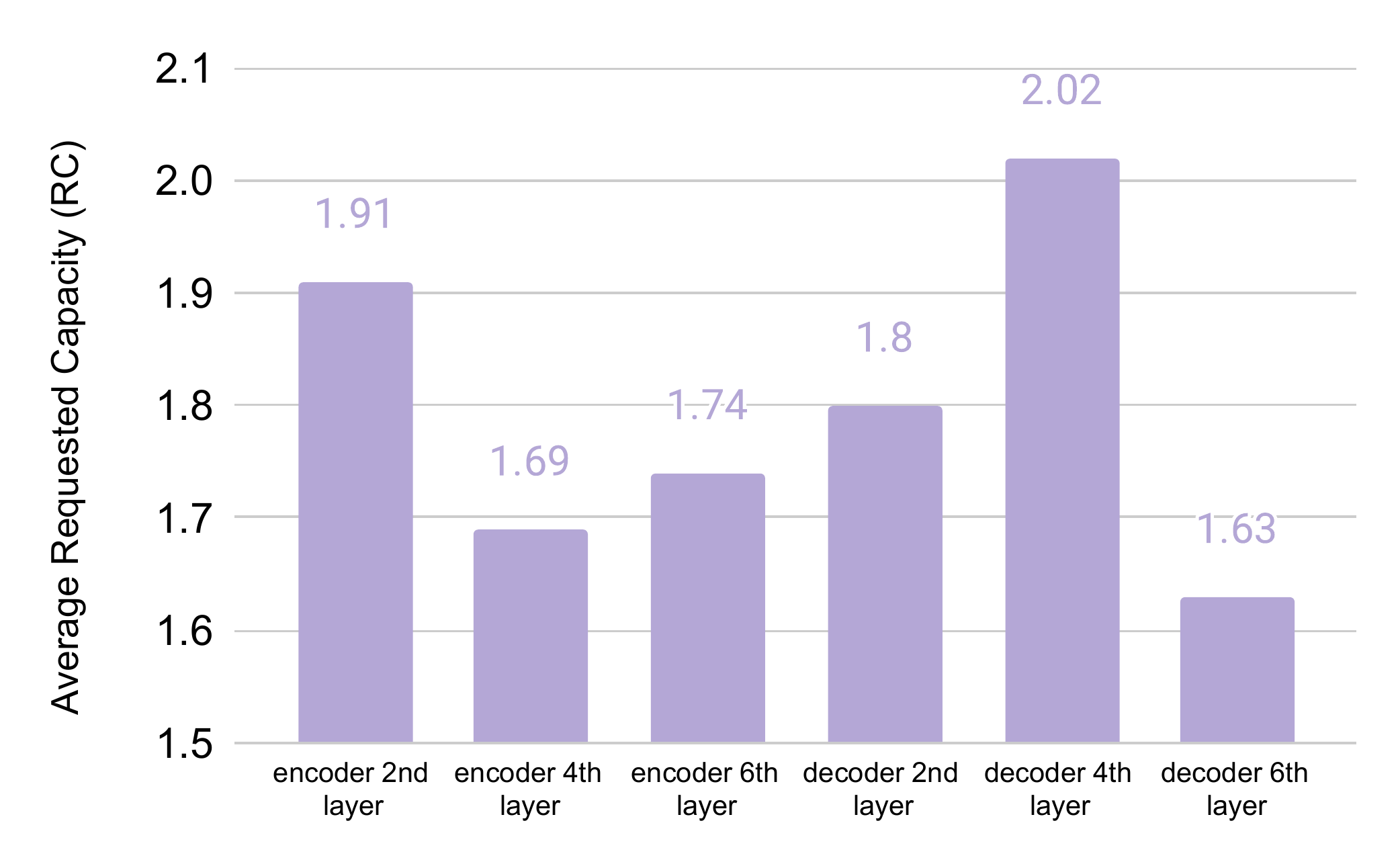}}
    \caption{Average \rc of \smoe blocks in different locations of the architecture.
    }
    \label{fig:depth}
\end{figure}

We analyze in this section the average \rc in relation to the location of the \smoe block in the transformer architecture.
As depicted in Figure \ref{fig:depth}, \rc varies depending on the location of the \smoe block.
Early encoder layers (encoder 2nd layer is the first \smoe block in the model) request more capacity than the subsequent encoder \smoe blocks. We hypothesize that this first block takes on the task of mapping tokens coming from different languages and different scripts to a shared space.
% On the decoder side, the middle decoder layer requests more capacity than other decoder layers.
% indicating that the \rc is influenced by the position of the \smoe blocks in the model.

\section{Conclusion}
This work presents Stratified Mixture of Experts (\smoe) models, a novel design for MoEs that is capable of dynamically assigning capacity to input tokens.
Through experimental evaluation at three scales (M4, M15, and OPUS-100), we have demonstrated that the proposed \smoe model surpasses the performance of many current state-of-the-art MoE methods.
This proves that dynamically assigning capacity to tokens in MoE models is a viable solution to address the MoE's parameter inefficiency.
% , raising the question of the efficiency of the constant capacity utilization employed in vanilla MoE models, which may contribute to their parameter inefficiency.
Additionally, we conduct a thorough analysis to investigate the factors that influence dynamic capacity assignment, including the language of the tokens and the location of the \smoe block within the model architecture.

\section*{Limitations}
Stratified Mixture of Experts (\smoe) aims to improve the performance of Mixture of Experts (MoE) models by assigning dynamic capacity to different tokens. While \smoe has demonstrated performance improvements over many state-of-the-art baselines, it also comes with an additional computational cost compared to traditional MoE models. 
% This is because \smoe consists of multiple \strata, each containing a set of experts, and tokens may pass through a variable number of these experts, depending on the routing decisions made by the gates at each \stratum. This dynamic routing mechanism introduces additional computation as indicated in Section \ref{sec:compute_cost}. 
However, the cost is small and the benefits of \smoe in terms of improved performance often outweigh this added computational cost, especially in tasks where performance is critical. For example, in OPUS-100, with 8\% FLOPs/tok, \config{4-12} achives +1.01 BLEU compared with traditional MoE \citep{gshard}.

% \section*{Ethics Statement}
% Scientific work published at ACL 2023 must comply with the ACL Ethics Policy.\footnote{\url{https://www.aclweb.org/portal/content/acl-code-ethics}} We encourage all authors to include an explicit ethics statement on the broader impact of the work, or other ethical considerations after the conclusion but before the references. The ethics statement will not count toward the page limit (8 pages for long, 4 pages for short papers).

\section*{Acknowledgements}
We thank anonymous reviewers for their insightful feedback. We also extend our gratitude to Steven Tan and Yunmo Chen for their valuable suggestions.

\bibliography{custom}
\bibliographystyle{acl_natbib}
\clearpage

\appendix
\begin{table*}[ht]
\centering
\resizebox{1\linewidth}{!}{
\begin{tabular}{lcwr{2cm}@{\hspace{2pt}}wl{2em}ccc}
\hline
Language           & \multicolumn{1}{c}{Language code} & \multicolumn{2}{c}{Parallel Data Size} & Resource Level & \multicolumn{1}{c}{Language family} \\ \hline
Northern Sotho     & \texttt{nso}                             & 526 & K                                   & Low          & Central Narrow Bantu                                              \\
Rundi              & \texttt{run}                             & 454 & K                                   & Low          & Central Narrow Bantu                                             \\
Swati              & \texttt{ssw}                             & 94 & K                                    & Very Low     & Central Narrow Bantu                                            \\
Indonesian         & \texttt{ind}                             & 6.5 & M                                   & High         & Malayo-Polynesian                                                      \\
Malay              & \texttt{msa}                             & 1 & M                                     & High         & Malayo-Polynesian                                                      \\
Tagalog            & \texttt{tgl}                             & 1 & M                                     & High         & Malayo-Polinesian                 &                                \\
Bokmål (Norwegian) & \texttt{nob}                             & 238 & K                                   & Low          & North Germanic                                                    \\
Icelandic          & \texttt{isl}                             & 1 & M                                     & High         & North Germanic                                                        \\
Faroese            & \texttt{fao}                             & 4 & K                                     & Very Low     & North Germanic                                                         \\
Slovene            & \texttt{slv}                             & 15 & M                                    & High         & Southwestern Slavic                                                     \\
Luxembourgish      & \texttt{ltz}                             & 8 & K                                     & Very Low     & Western Germanic                                                       \\
Limburgish         & \texttt{lim}                             & 5 & K                                     & Very Low     & Western Germanic                                                        \\
Catalan            & \texttt{cat}                             & 634 & K                                   & Low          & Western Romance                                                       \\
Galician           & \texttt{glg}                             & 195 & K                                   & Low          & Western Romance                                                          \\
Friulian           & \texttt{fur}                             & 6 & K                                     & Very Low     & Western Romance                                                        \\ \hline
\end{tabular}
}
\caption{Information about the 15 languages in the M15 dataset.}
\label{tab:m15_info}
\end{table*}
\section{M15 Information}
\label{app:m15_info}
We show the detailed information of the M15 dataset in Table \ref{tab:m15_info}.

\section{Additional Training Details}
\label{app:train_detail}
We employ the transformer$_\texttt{base}$ model (with an FFN dimension of 2048 and an embedding dimension of 512) for the M4 dataset, and the transformer$_\texttt{big}$ model (with an FFN dimension of 4096 and an embedding dimension of 1024) for M15 and OPUS-100 dataset. The maximum learning rate is 0.0008 for M4 and M15, and 0.0005 for the OPUS-100 dataset. The optimizer is Adam \citep{kingma2014adam} with \texttt{inverse\_sqrt} learning rate scheduler and weight decay of 0. The total number of training steps is 100K with 8K warm-up steps. The batch size is 13K tokens for M4 and M15, and 65K tokens for OPUS-100.

% \section{High-Frequent Tokens with Low NoC}
% \label{app:high_freq_low_noc}
% It is counter-intuitive that high-frequent tokens are inclined to have the least Requested Capacity (NoC). However, we notice that most of them are basic  
\end{document}